\documentclass[review]{elsarticle}

\usepackage{amsfonts}
\usepackage{amsmath}
\usepackage{graphicx,setspace,algorithmic,algorithm,multirow,cases,amssymb,subfigure,setspace}
\usepackage{lettrine}
\usepackage{multirow}
\usepackage{xcolor}
\usepackage{colortbl}
\usepackage{times}
\usepackage{graphicx}
\usepackage{amsmath}
\usepackage{indentfirst}
\usepackage{booktabs}
\usepackage{blindtext}
\usepackage{array}
\usepackage{tabularx,booktabs}
\usepackage{color, soul}
\usepackage{bm}
\usepackage{array}
\usepackage{booktabs}
\usepackage{pifont}
\usepackage{hyperref}

\journal{Journal of Neurocomputing}

\begin{document}

\begin{frontmatter}
\title{Representation Discrepancy Bridging Method for Remote Sensing Image-Text Retrieval}

\author[a,b,c]{Hailong Ning}
\author[a,b,c]{Siying Wang}
\author[e]{Tao Lei}
\cortext[d]{Corresponding author}
\author[a,b,c]{Xiaopeng Cao}
\author[f]{Huanmin Dou}
\author[g]{Bin Zhao}
\author[h]{Asoke K. Nandi}
\author[i]{Petia Radeva\corref{d}}

\address[a]{School of Computer Science and Technology, Xi’an University of Posts and Telecommunications, Xi’an 710121, China.}
\address[b]{Shaanxi Key Laboratory of Network Data Analysis and Intelligent Processing, Xi’an 710121, China.}
\address[c]{Xi’an Key Laboratory of Big Data and Intelligent Computing, Xi’an 710121, China.}
\address[e]{Shaanxi Joint Laboratory of Artificial Intelli gence and the School of Electronic Information and Artificial Intelligence, Shaanxi University of Science and Technology, Xi’an 710021, China.}
\address[f]{School of Computer Engineering, Weifang University, Shandong 261061, China.}
\address[g]{Shanghai Artificial Intelligence Laboratory, Shanghai, 200003, China.}
\address[h]{Department of Electronic and Electrical Engineering, Brunel University of London, Uxbridge, UB8 3PH, United Kingdom.}
\address[i]{Dept. Matemàtiques i Informàtica and Institute of Neuroscience, Univeritat de Barcelona, Barcelona, 08007, Spain.}

\begin{abstract}
Remote Sensing Image-Text Retrieval (RSITR) plays a critical role in geographic information interpretation, disaster monitoring, and urban planning by establishing semantic associations between image and textual descriptions. Existing Parameter-Efficient Fine-Tuning (PEFT) methods for Vision-and-Language Pre-training (VLP) models typically adopt symmetric adapter structures for exploring cross-modal correlations. However, the strong discriminative nature of text modality may dominate the optimization process and inhibits image representation learning. The nonnegligible imbalanced cross-modal optimization remains a bottleneck to enhancing the model performance. To address this issue, this study proposes a Representation Discrepancy Bridging (RDB) method for the RSITR task. On the one hand, a Cross-Modal Asymmetric Adapter (CMAA) is designed to enable modality-specific optimization and improve feature alignment. The CMAA comprises a Visual Enhancement Adapter (VEA) and a Text Semantic Adapter (TSA). VEA mines fine-grained image features by Differential Attention (DA) mechanism, while TSA identifies key textual semantics through Hierarchical Attention (HA) mechanism. On the other hand, this study extends the traditional single-task retrieval framework to a dual-task optimization framework and develops a Dual-Task Consistency Loss (DTCL). The DTCL improves cross-modal alignment robustness through an adaptive weighted combination of cross-modal, classification, and exponential moving average consistency constraints. Experiments on RSICD and RSITMD datasets show that the proposed RDB method achieves a 6\%-11\% improvement in mR metrics compared to state-of-the-art PEFT methods and a 1.15\%-2\% improvement over the full fine-tuned GeoRSCLIP model.
\end{abstract}

\begin{keyword}
Remote Sensing Image-Text Retrieval, Vision-and-Language Pre-training, Parameter-Efficient Fine-Tuning, Cross-Modal Asymmetric Adapter, Multi-Task Learning
\end{keyword}

\end{frontmatter}


\section{INTRODUCTION}
Remote Sensing Image-Text Retrieval (RSITR) aims to align satellite images with textual descriptions by bridging semantic associations between heterogeneous modalities. It has emerged as a critical technology for various earth observation applications such as geographic information interpretation \cite{jovhari2025noise}, disaster monitoring \cite{WANG2024109501}, and urban planning \cite{DONG2025100186}. The primary challenge for RSITR is to bridge the modality gap and construct a robust cross-modal semantic alignment model \cite{ZHANG9540028,NING9371014}.

Currently, the mainstream RSITR methods can be broadly divided into naive deep learning based \cite{rs12030405,HOXHA9154525,app10248931} and Vision-Language Pretraining (VLP) based methods \cite{li2025benchmark,LU2024}. Naive deep learning based methods usually employ various deep learning operators such as Convolutional Neural Network (CNN), Recurrent Neural Network (RNN), and Transformer to extract features from different modalites. Subsequently, these features are projected into a shared feature space, where cross-modal matching is performed using similarity metric. However, such methods often struggle to accurately align features due to the limited generalization performance when confronted with the complex cross-modal relationships between image and text modalites. To address this limitation, researchers have increasingly adopted domain-adaptive fine-tuning strategies based on large-scale VLP models \cite{KHATTAK2023,YUAN2023}. By keeping the backbone network parameters fixed and adjusting only a small subset of task-specific parameters, such methods can preserve general semantic knowledge while incorporating domain-specific characteristics. This not only significantly reduces training cost, but also enhances the cross-modal matching accuracy in RS scenarios.

In recent years, VLP models have demonstrated notable advances in cross-modal learning. The most representative one is the Contrastive Language-Image Pretraining (CLIP) \cite{pmlr-v139-radford21a} model, which aims to map image and text modalities into a shared multi-modal embedding space for semantic alignment through contrast learning. During training, CLIP utilizes a contrastive loss to bring the embeddings of aligned image-text pairs closer while pushing apart the embeddings of mismatched pairs. Based on CLIP, researchers have successively developed pre-trained models such as BLIP \cite{pmlr-v162-li22n}, RemoteCLIP \cite{LIU2024} and GeoRSCLIP \cite{ZHANG10679571}. BLIP constructs a multimodal hybrid architecture to improve image-text comprehension and generation capabilities. RemoteCLIP and GeoRSCLIP are specifically tailored for RS data to improve model generalization in the RS domain. In fact, RS images exhibit distinct domain characteristics compared to natural images, as they are typically captured by high-altitude or space-based platforms, such as satellites and drones. In addition, RS images often contain multiple objects and exhibit high inter-class diversity. These domain-specific differences makes it difficult to directly transfer knowledge learned from natural scenes to the RS domain. Therefore, efficiently performing domain-adaptive Fine-Tuning (FT) on large-scale VLP models becomes the key to mitigate the domain difference between natural and RS scenes.

\begin{figure}[tp]
\begin{center}
\includegraphics[width=1.0\linewidth]{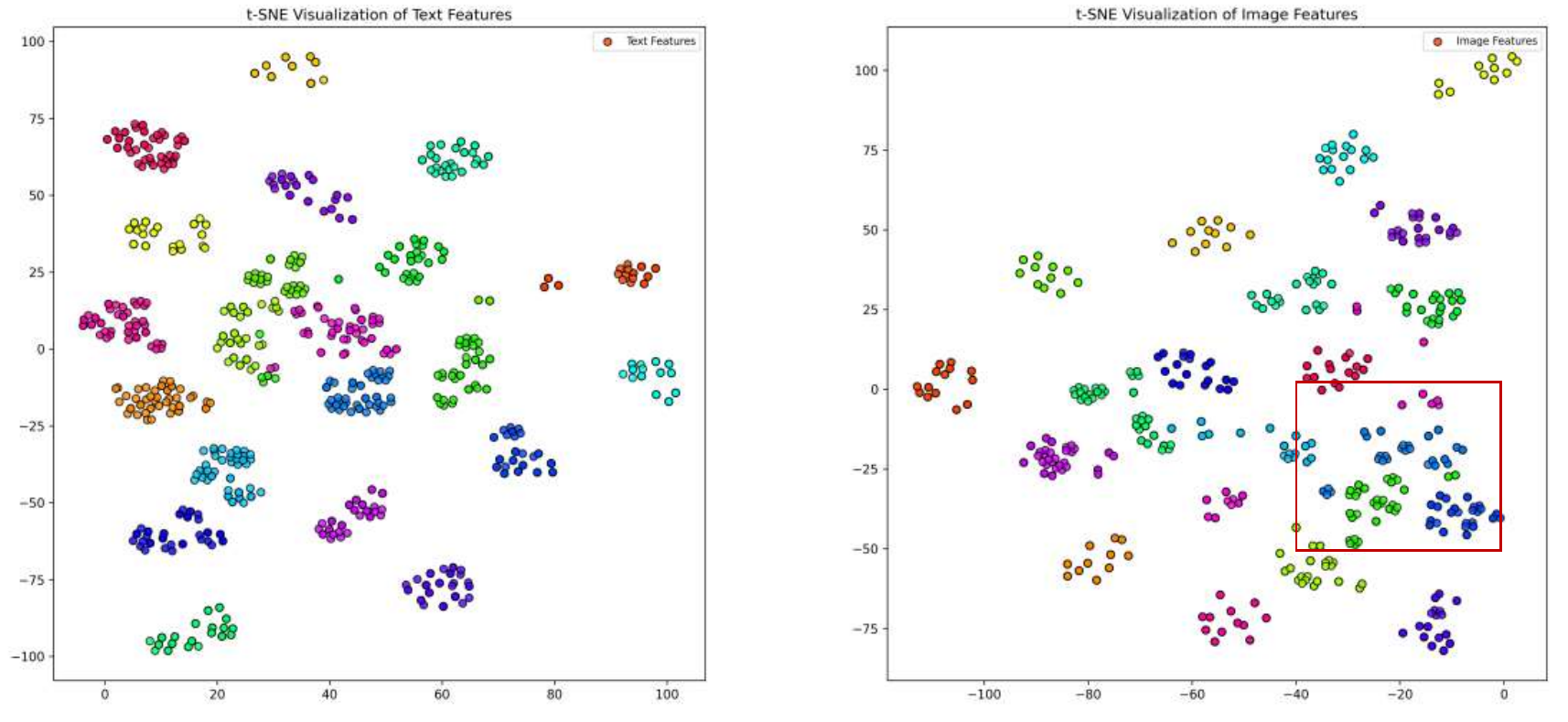}
\renewcommand{\figurename}{Fig.}
\end{center}
    \caption{\small{The t-SNE visualization of features from text modality (left panel) and image modality (right panel) when utilizing a symmetric adapter structure for the RSITR task.}}
\label{fig:1}
\end{figure}

Traditional full parameter fine-tuning methods have demonstrated superior performance in RSITR tasks. However, these methods suffer from substantial training costs and are prone to catastrophic forgetting. Practically, the number of fine-tuning parameters directly impacts the adaptive ability and training efficiency of models. Consequently, the Parameter-Efficient Fine-Tuning (PEFT) method has gained widespread attention. The main concept of PEFT is to keep the majority of the pretrained parameters unchanged while only fine-tuning a limited number of newly added parameters tailored to the target task. This strategy effectively preserves the general knowledge of original models while integrating domain-specific information efficiently. In PEFT methods, adapter \cite{pmlr-v97-houlsby19a} is the most commonly adopted fine-tuning scheme. However, in the RSITR task, the structural design of adapter typically employs symmetric architecture for both image and text modalities. This design ignores the inherent differences in representation learning between modalities, potentially limiting the model to capture modality-specific features. Notably, image modality poses greater challenges for key information extraction due to its inherent complexity and redundant content. In contrast, textual description provides concise information with more explicit key elements. Figure \ref{fig:1} illustrates the t-SNE visualization results of features from both modalities when using a symmetric adapter structure for the RSITR task. It can be seen that the text modality features (left panel) exhibit significantly higher inter-class discriminability compared to the image modality (right panel), particularly in the area marked by the red rectangle. The higher inter-class discriminability of text modality may dominate the cross-modal optimization process and inhibits image representation learning. Motivated by these observations, this study argues that an efficient asymmetric adapter should be explored to deal with the imbalanced cross-modal optimization problem.

To this end, this study proposes a Representation Discrepancy Bridging (RDB) method for the RSITR task. For the network architecture, a Cross-Modal Asymmetric Adapter (CMAA) structure is designed, comprising an asymmetric Visual Enhancement Adapter (VEA) and Text Semantic Adapter (TSA). The VEA leverages a Differential Attention (DA) mechanism \cite{ye2025differential} to extract fine-grained image features, while the TSA employs a Hierarchical Attention (HA) mechanism \cite{yang-etal-2016-hierarchical} to identify key textual semantics. For the model optimization, a Dual-Task Consistency Loss (DTCL) function is proposed to enhance inter-modal alignment robustness through the joint optimization of cross-modal semantic alignment and intra-modal semantic discrimination.

In summary, the main contributions of this study are threefold:
\begin{itemize}
  \item A Cross-Modal Asymmetric Adapter (CMAA) structure is proposed to solve the imbalanced cross-modal optimization problem. Different from the existing symmetric adapter structure, CMAA employs asymmetric Visual Enhancement Adapter (VEA) and Text Semantic Adapter (TSA) for modality-specific fine-tuning.
  \item A Dual-Task Consistency Loss (DTCL) function is designed to enhance cross-modal alignment robustness. Unlike existing cross-modal retrieval models based on single-task optimization, the DTCL loss adopts a dual-task optimization framework for facilitating cross-modality alignment by joint optimization of cross-modal semantic alignment and intra-modal semantic discrimination.
  \item Based on the above design, a Representation Discrepancy Bridging (RDB) method is proposed for efficient RSITR. It achieves dual improvements by introducing asymmetric adapter fine-tuning and dual-task consistency learning. Extensive experiments demonstrate that the proposed RDB method improves the retrieval performance by 1.15\%-2\% compared to the full fine-tuned GeoRSCLIP.
\end{itemize}

\section{RELATED WORK}\label{RELATEDWORK}
\subsection{Remote Sensing Image-Text Retrieval Methods}\label{RemoteSensingImage-TextRetrievalMethods}
As RS vision-language technology continues to advance, RSITR has become an important research hotspot in geospatial intelligence applications. However, due to the inherent heterogeneity and modality-specific characteristics of RS images and textual descriptions, current RSITR methods \cite{rs12030405,HOXHA9154525,app10248931,CHENG2021} mainly employ deep learning techniques for automatic feature extraction. For instance, Hoxha {\it et al.} \cite{HOXHA9154525} proposed a Remote Sensing Image Retrieval (RSIR) method that encodes visual features of RS images into textual descriptions (i.e., captions) and retrieves the most similar archived image by calculating the similarity between their feature vectors. Rahhal {\it et al.} \cite{app10248931} introduced an unsupervised method for retrieving text-image pairs in RS images, which uses the Big Transfer (BiT) model and Bi-directional Long-Short Term Memory (Bi-LSTM) network to extract features. In addition, it designs an embedding loss to align image-text pairs while preserving inter-sample differences. In order to mine the content of RS images more comprehensively, Lu {\it et al.} \cite{LU2018} constructed an aerial image dataset RSICD. To address the overfitting problem on the RSICD dataset, Li {\it et al.} \cite{LIX2021} introduced a novel TCE loss function. It mitigates the overfitting effect caused by the traditional Cross-Entropy (CE) loss. Wang {\it et al.} \cite{WANG2019} employed a Multimodal Tensor Fusion Network (MTFN). The image-text similarity function is explicitly learned through a rank-decomposition based fusion model, thus balancing both retrieval accuracy and model complexity. Besides, the quality of retrieved images is also very important for RSITR. Li {\it et al.} \cite{li2016spatiotemporal} proposed a quality assessment method. It is an impressive work that achieves the intersection of the imaging process and processing process and has a profound impact on the development of image processing.

In response to computational efficiency demands, Yuan {\it et al.} \cite{YUANZ2022} proposed a lightweight and efficient cross-modal retrieval model LW-MCR. The model employs lightw-eight group convolution for feature extraction and enhances retrieval performance through implicitly supervised knowledge distillation and comparative learning of unlabeled data. In addition, Yuan {\it et al.} \cite{YUAN20222} developed a more detailed and demanding RS Image-Text Match Dataset (RSITMD). Subsequently, they \cite{YUAN20222} proposed the Multi-level Information Dynamic Fusion (MIDF) module, the Denoised Representation matrix and the Enhanced Adjacency matrix (DREA). They also introduced the Multivariate Rerank (MR) algorithms to address the problem of neglecting the target relationships and salient local features. Cheng {\it et al.} \cite{CHENG2021} proposed a cross-modal retrieval network to establish a direct association between RS images and paired texts. The network introduces a semantic alignment module and combines attention and gating mechanisms to learn discriminative features. Recently, Liu {\it et al.} \cite{LIU2024} introduced RemoteCLIP, the first Vision-Language foundation model specifically designed for RSITR. The model aims to learn semantically rich visual features and aligns text embeddings through contrastive learning. Rahhal {\it et al.} \cite{RAHHAL2022} developed a Transformer-based multilingual cross-modal RSITR framework leveraging pretrained CLIP models, which significantly outperforms traditional deep learning methods.

\subsection{Vision-Language Pre-training Models and Their Fine-Tuning}\label{Vision-LanguagePre-trainingModelsandTheirFine-Tuning}
In recent years, large-scale VLP models have made notable progress. Based on their encoder design, VLP models are generally divided into fusion encoder and dual encoder types. The fusion encoder is further divided into single-stream architecture and dual-stream architecture. Single-stream architectures (e.g., SimVLM \cite{wang2022simvlm}, M6 \cite{lin2021m6}, ViLT \cite{pmlr-v139-kim21k}, etc.) concatenate textual and visual features before processing them through a unified Transformer module. However, this design tends to ignore deep inter-modal interactions due to the direct application of self-attention operation for concatenated features. To address this limitation, dual-stream architectures (e.g., ViLBERT \cite{LUJ2019}, ERNIL-ViL \cite{YU2021}, TDN \cite{WANG2021}, etc.) employ separate modules to independently process text and image features, which enables cross-modal interaction through cross-attention mechanisms. Notably, the aforementioned fusion encoders often encounter computational bottlenecks in RSITR tasks due to exhaustive pairwise image-text encoding requirements. In contrast, dual encoders (e.g., CLIP \cite{pmlr-v139-radford21a}, RemoteCLIP \cite{LIU2024}, GeoRSCLIP \cite{ZHANG10679571}, etc.) demonstrate superior scalability through modality-independent encoding and joint semantic subspace projection, making them particularly suitable for large-scale RS applications. For example, RemoteCLIP and GeoRSCLIP leverage RS-tailored pre-training strategies to attain state-of-the-art generalization results using efficient dual-encoder frameworks. More recently, Li {\it et al.} \cite{li2022positive} developed a novel paradigm for multimodal alignment and fusion, which first scientifically investigates the positive effects of noise, and pioneers a series of research on noise utilization in VLP models. 

Large-scale VLP model fine-tuning utilizes domain-specific data to optimize pre-trained models. This improves their performance in specific tasks. The main fine-tuning methods include Full fine-tuning and PEFT. Full fine-tuning utilizes the full potential of a pre-trained model by tuning all its parameters. However, it is computationally intensive and requires significant training time. In contrast, PEFT achieves efficient domain adaptation with minimal resources by reducing the number of fine-tuning parameters and computational overhead. PEFT primarily consists of prompt tuning \cite{LESTER2021}, prefix tuning \cite{LI2021} and adapter tuning \cite{pmlr-v97-houlsby19a}. Among them, prompt tuning guides the model to perform a task by incorporating “prompts” into the input layer. For example, Jia {\it et al.} \cite{JIA2022} proposed Visual Prompt Tuning (VPT), which introduces trainable visual prompts while keeping the main model parameters frozen, providing an efficient alternative to fine-tuning. Khattak {\it et al.} \cite{KHATTAK2023} developed Multi-modal Prompt Learning (MaPLe), which designs coupled vision-language cues and progressively models feature relationships to enable richer contextual learning. In comparison, prefix tuning achieves more flexible and efficient optimization by adding learnable prefixes to the input and performing layer-wise weighting to help the model extract task-specific features. For instance, Wu {\it et al.} \cite{WU2024} proposed Adaptive Prefix Tuning (APT) by using a gating mechanism to dynamically optimize prefixes for learning layer-specific representations. However, both prompt tuning and prefix tuning mainly adapt to the task by designing specific input structures, which limits their capacity for multi-task learning. As a result, adapter tuning has become one of the most commonly used methods within PEFT. It achieves efficient fine-tuning without modifying the original model parameters by inserting Adapter modules with a small number of parameters between specific layers of the pre-trained model. Gao {\it et al.} \cite{GAO2023} proposed CLIP-Adapter, which enhances VLP model fine-tuning by introducing feature adapters instead of using traditional Prompt Tuning methods. Recently, Yuan {\it et al.} \cite{YUAN2023} introduced a Parameter-Efficient Transfer Learning (PETL) framework that facilitates the transfer of vision-language knowledge from the natural image domain to remote sensing, and presented MRS-Adapter to improve multi-modal fusion for RSITR tasks. Generally, Adapter tuning has gained popularity in PEFT by optimizing a small number of plug-and-play modules.

Although current PEFT methods have significantly improved the performance of RSITR models, they typically employ symmetric adapter architectures. This fails to fully considering the inherent differences between image and text modality features, which may lead to imbalanced cross-modal optimization problem. To address this limitation, a Cross-Modal Asymmetric Adapter (CMAA) architecture is designed to facilitate modality-specific optimization and improve feature alignment. Meanwhile, the DTCL loss function is proposed to extend the traditional single-task retrieval model into a multi-task optimization model for effectively improving the RSITR model performance.

\begin{figure*}[tp]
\begin{center}
\includegraphics[width=1.2\linewidth]{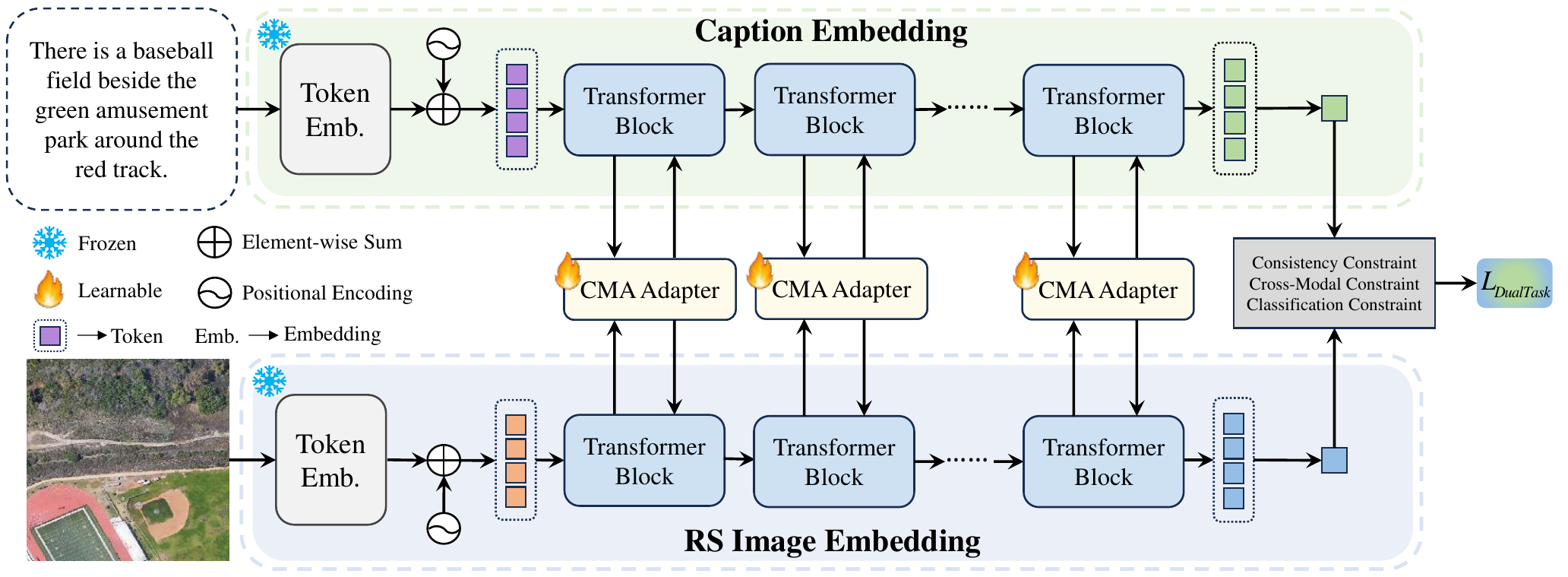}
\renewcommand{\figurename}{Fig.}
\end{center}
    \caption{\small{The overall framework of the RDB method.}}
\label{fig:overall}
\end{figure*}

\section{METHODOLOGY}

\subsection{Overall Framework}

The overall framework of the proposed Representation Discrepancy Bridging (RDB) method is shown in Figure~\ref{fig:overall}. The method employs ViT as the image encoder and BERT as the text encoder. Similar to the CLIP model, both encoders consist of 12 Transformer blocks for feature extraction. Specifically, the image is first fed into the image encoder and subsequently undergoes feature processing through the Visual Enhancement Adapter (VEA). At the same time, the text is fed into the text encoder and further refined using the Text Semantic Adapter (TSA). Thus, the corresponding key features are extracted respectively according to the characteristics of each modalities. In addition, to enhance the interaction between the two modalities, a shared layer is incorporated in the Cross-Modal Asymmetric Adapter (CMAA) to facilitate cross-modal complementarity and strengthen inter-modal connections. Finally, the Dual-Task Consistency Loss (DTCL) loss function is introduced to optimize and strengthen the inter-modal consistency learning, thereby improving the performance of the RSITR task.


\subsection{Cross-Modal Asymmetric Adapter}
Recent years, adapters are introduced to the pre-trained models, with the goal of handling new tasks by fine-tuning only a few parameters. A standard adapter is composed of multiple linear layers and nonlinear activation functions defined as:
\begin{equation}
    \mathrm{Adapter}(\mathbf{E})=[\mathbf{W}^\mathrm{up}\sigma(\mathbf{W}^\mathrm{down}\mathbf{E}^\mathrm{T})]^\mathrm{T}
\end{equation}
where $\mathbf{E}$ is the input feature for adapters, $\mathbf{W}^{\mathrm{up}}$
 and $\mathbf{W}^\mathrm{down}$ represent the weights for the down-projection and up-projection respectively, and $\sigma$ is the nonlinear activation function.

In the RSITR task, existing methods typically design adapters with the symmetric architecture for both image and text modalities. However, due to the significant representation discrepancy between image and text data, simply applying symmetric adapter for the two modalities often fails to effectively capture modality-specific features. This limitation significantly impacts the overall performance of RSITR models.

\begin{figure*}[tp]
\begin{center}
\includegraphics[width=1.0\linewidth]{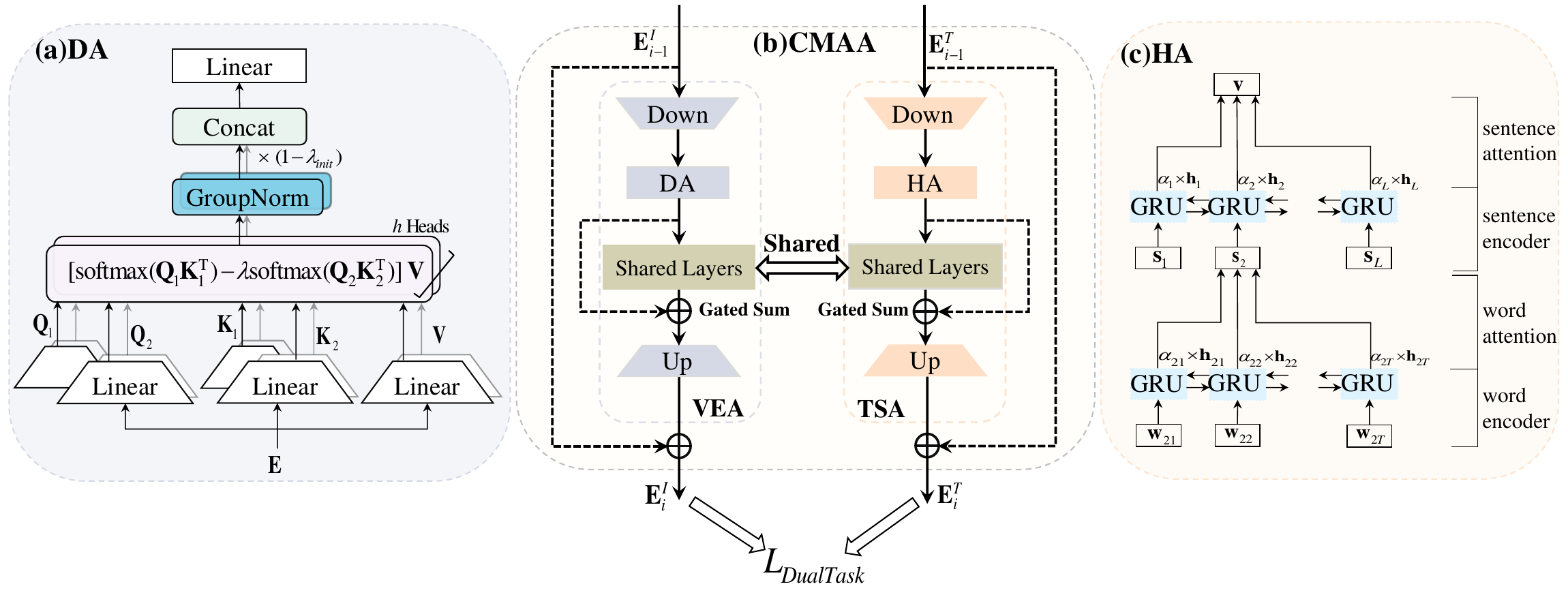}
\renewcommand{\figurename}{Fig.}
\end{center}
    \caption{\small{(a) The DA mechanism introduced in the VEA. (b) The specific structure of the CMAA. (c) The HA mechanism introduced in the TSA.}}
\label{fig:adapter}
\end{figure*}

To alleviate the representation discrepancy between RS image and text modalities, this study proposes a Cross-Modal Asymmetric Adapter (CMAA). As illustrated in Figure~\ref{fig:adapter}(b), the CMAA incorporates Visual Enhancement Adapter (VEA) and Text Semantic Adapter (TSA) to optimize the modality-specific features for the inherent characteristics of image and text modalities, respectively.

As shown in Figure~\ref{fig:adapter}, VEA incorporates the DA mechanism to effectively mine fine-grained features from the image modality. This mechanism dynamically adjusts the attention scale based on task requirements, allowing for more precise feature extraction. Concurrently, TSA introduces the HA mechanism to emphasize semantically discriminative linguistic elements. By dynamically identifying task-critical keywords while suppressing irrelevant stop words and auxiliary components, HIA ensures efficient textual feature learning. In the CMAA module, the feature extraction pipeline operates through three sequential phases. First, the image feature $\mathbf{E}^{I}$ and text feature $\mathbf{E}^{T}$ through Transformer blocks undergo a dimensionality reduction via linear projection, followed by GELU activation function to introduce non-linearity. Subsequently, the image modality is processed by DA and the text modality is processed by HA to further enhance the feature representations. Then, these features are fed into the shared layer for inter-modal interaction. Finally, an up-projection layer reconstructs the shared representations to match the original input dimensions, preserving compatibility with subsequent processing stages.

\subsubsection{Visual Enhancement Adapter}
In remote sensing imagery, fine-grained feature extraction is critical due to high intra-class similarity and significant redundant information. For this reason, the DA mechanism is introduced in VEA (shown in Figure~\ref{fig:adapter}(a)). It enhances the fine-grained representation by dynamically capturing the relationship between local details and the global context. Specifically, the DA mechanism transforms the query, key and value vectors into output representations. It computes attention scores using the query and key vectors, which are then weighted and aggregated over the value vectors. The DA mechanism can be expressed as:
\begin{equation}
    \begin{aligned}&[\mathbf{Q}_{1};\mathbf{Q}_{2}]=\mathbf{E}\mathbf{W}^{Q},\quad[\mathbf{K}_{1};\mathbf{K}_{2}]=\mathbf{E}\mathbf{W}^{K},\quad\mathbf{V}=\mathbf{E}\mathbf{W}^{V}\\&\mathrm{DiffAttn}(\mathbf{E})=(\mathrm{softmax}(\frac{\mathbf{Q}_1\mathbf{K}_1^\mathrm{T}}{\sqrt{d}})-\lambda\mathrm{~softmax}(\frac{\mathbf{Q}_2\mathbf{K}_2^\mathrm{T}}{\sqrt{d}}))\mathbf{V}
    \end{aligned}
\end{equation}
where $\mathbf{Q}$, $\mathbf{K}$, $\mathbf{V}$ is the query, key and value vectors, respectively. Both $\mathbf{Q}$ and $\mathbf{K}$ are split into two components $\mathbf{Q}_{1}$, $\mathbf{Q}_{2}$ and $\mathbf{K}_{1}$, $\mathbf{K}_{2}$ for subsequent difference computation. $\mathbf{W}^{Q}$, $\mathbf{W}^{K}$, and $\mathbf{W}^{V}$ are the parameters for obtaining $\mathbf{Q}$, $\mathbf{K}$, and $\mathbf{V}$, respectively. $\lambda$ is a learnable scalar. To coordinate the dynamics of the learning process, $\lambda$ is reparameterized as:
\begin{equation}
    \lambda=\exp(\lambda_{q_1}\cdot\lambda_{k_1})-\exp(\lambda_{q_2}\cdot\lambda_{k_2})+\lambda_{init}
\end{equation}
where $\lambda_{q_1}$, $\lambda_{k_1}$, $\lambda_{q_2}$, and $\lambda_{k_2}$ are learnable vectors; $\lambda_{init}\in(0,1)$ is a constant for initializing $\lambda$.

The features extracted by the DA mechanism are subsequently forwarded to a shared layer that enables information sharing with the text modality, thereby reinforcing the complementarity between the two modalities. For example, when key information in the text is omitted or inadequately expressed, the visual cues from the image can be used as a supplement for enhancing the comprehension and reasoning capabilities of models. Finally, a gating mechanism is introduced to regulate the interaction between different modalities. This mechanism can prevent any single modality from dominating or being neglected.

\subsubsection{Text Semantic Adapter}
In remote sensing scenarios, textual descriptions are often abstract or generalized, which makes it challenging to explicitly convey the fine-grained details of remote sensing images. Therefore, this study introduces the HA mechanism to identify and extract key information words (as shown in Figure~\ref{fig:adapter}(c)). The extracted representations are subsequently aggregated to construct sentence vectors. Specifically, the words $\mathbf{w}$ are first encoded by a bi-directional GRU to obtain word-level annotations $\mathbf{h}_{it}$. Then, $\mathbf{h}_{it}$ is fed into a MLP to obtain their hidden representations $\mathbf{u}_{it}$. Next, the importance of each word is determined by the measuring the similarity between its hidden representation $\mathbf{u}_{it}$ and the word-level context vector $\mathbf{u}_{w}$. The resulting scores are then normalized using a softmax function to obtain importance weights $\alpha_{it}$. Finally, the word-level annotations are weighted and summed according to these normalized weights to obtain sentence vectors $\mathbf{s}_{i}$. Notably, the word-level context vector $\mathbf{u}_{w}$ can be regarded as a high-level representation of a fixed query “which word is the most informative word”, similar to the query mechanism in Memory Networks\cite{kumar, Sukhbaatar}. The HA mechanism can be formally expressed as follows:
\begin{align}
    &\mathbf{u}_{it}=\tanh(\mathbf{W}_w\mathbf{h}_{it}+\mathbf{b}_w) \\
    &\alpha_{it}=\frac{\exp(\mathbf{u}_{it}^\mathrm{T}\mathbf{u}_w)}{\sum_t\exp(\mathbf{u}_{it}^\mathrm{T}\mathbf{u}_w)} \\
    &\mathbf{s}_i=\sum_t\alpha_{it}\mathbf{h}_{it}
\end{align}
Subsequently, the sentence vectors $\mathbf{s}_{i}$ is encoded by a bi-directional GRU to obtain sentence-level annotations $\mathbf{h}_{i}$. Then a sentence-level context vector $\mathbf{u}_{s}$ is introduced to the sentence vectors to measure the importance of each sentences. Similar to the word-level processing method, this study adopts the same strategy to obtain the hidden representation $\mathbf{u}_{i}$ of sentence vectors, the sentence-level importance weight $\alpha_{i}$, and the text vector $\mathbf{v}$ for integrating all the sentence information. This process is formally defined as follows:
\begin{align}
    &\mathbf{u}_i=\tanh(\mathbf{W}_s\mathbf{h}_i+\mathbf{b}_s) \\
    &\alpha_i=\frac{\exp(\mathbf{u}_i^\mathrm{T}\mathbf{u}_s)}{\sum_i\exp(\mathbf{u}_i^\mathrm{T}\mathbf{u}_s)} \\
    &\mathbf{v}=\sum_i\alpha_i\mathbf{h}_i
\end{align}

After being processed by the DA mechanism and the HA mechanism, the subsequent operations of both VEA and TSA adopt up-projection layers to restore the feature size.


\subsection{DTCL Learning Objective}
To enhance cross-modal alignment robustness in RSITR, this study extends the traditional single-task retrieval model into a multi-task optimization framework and proposes Dual-Task Consistency Loss (DTCL). The DTCL improves feature representation robustness by jointly optimizing cross-modal semantic alignment and intra-modal semantic discrimination. DTCL contains three key constraint terms: cross-modal constraint, classification constraint, and exponential moving average consistency constraint.

(1) Cross-Modal Constraint: To achieve effective alignment between remote sensing images and textual descriptions, this study adopts the following loss function specifically designed to measure the semantic similarity between the two modalities:
\begin{align}
    L_{\mathrm{cross}}(\mathbf{I},\mathbf{T}) 
    = & \sum_{\hat{\mathbf{T}}}\left[\mu - \cos(\mathbf{I},\mathbf{T}) + \cos(\mathbf{I},\hat{\mathbf{T}})\right]_{+} \notag \\
    & + \sum_{\hat{\mathbf{I}}}\left[\mu - \cos(\mathbf{I},\mathbf{T}) + \cos(\hat{\mathbf{I}},\mathbf{T})\right]_{+}
\end{align}
where $(\mathbf{I},\mathbf{T})$ represents matched pairs of samples. $\hat{\mathbf{T}}$ denotes the text that does not match the RS image $\mathbf{I}$. $\hat{\mathbf{I}}$ denotes the RS image that does not match the text $\mathbf{T}$. $\mu$ is the boundary value for the cross-modal constraint. The cosine similarity is calculated as follows:
\begin{equation}
    \cos(\mathbf{I},\mathbf{T})=\frac{\sum_{i=1}^n\mathbf{I}_i\mathbf{T}_i}{\sqrt{\sum_{i=1}^n\mathbf{I}_i^2}\sqrt{\sum_{i=1}^n\mathbf{T}_i^2}}
\end{equation}

(2) Classification Constraint: To further enhance the semantic discriminative capability of each modality, a classification loss is introduced to measure the classification performance of the model. The loss quantifies the discrepancy between the predicted probability distribution and the true category distribution. The specific calculation formula as follows:
\begin{equation}
    L_{\mathrm{cls}}=-\sum_{c=1}^Cy_c\log(p_c)
\end{equation}
where $C$ represents the number of categories. $\mathbf{y}$ denotes the one-hot coding of the true label, where $y_c$ is the $c$-th element in $\mathbf{y}$. $p_c$ represents the predicted probability of the model for the $c$-th category.

(3) Consistency Constraint: To further alleviate the strong discriminability dominating the cross-modal optimization process, this study additionally introduces the Exponential Moving Average (EMA) mechanism to construct the teacher-student learning framework. In this framework, the text encoder plays the role of the teacher model and the image encoder plays the role of the student model. The text encoder guides the image encoder to generate more consistent sharing representation by minimizing the representation discrepancy between the two modalities. The specific calculation formula is as follows:
\begin{equation}
    L_{\mathrm{consist}}=\frac{1}{n}\sum_{i=1}^n\left(\mathbf{y}_{I_i}-\mathbf{y}_{T_i}\right)^2
\end{equation}
where $n$ is the number of samples, $\mathbf{y}_{I_i}$ denotes the output representation of the $i$-th sample image, and $\mathbf{y}_{T_i}$ is the output representation of the $i$-th sample text.

(4) Overall Objective: To integrate different loss terms, most of previous studies set different fixed weight parameters for each loss item. Although when the weight parameters are set appropriately, the model can be optimized very well. However, determining the appropriate weight parameters is tricky. Therefore, this study adopts an automatic weighted adjustment strategy \cite{LIEBEL2018} to solve this problem. Specifically, the three loss terms are dynamically weighted and optimized according to their relative contributions during training. This adaptive strategy enhances model learning and improves cross-modal feature alignment. By dynamically integrating cross-modal constraint, classification constraint, and exponential moving average consistency constraint, the DTCL loss function is defined as follows:
\begin{equation}
    L_{\mathrm{total}}=\sum_{i=1}^n\left(\frac{1}{2\sigma_i^2}L_i+\log(1+\sigma_i^2)\right)
\end{equation}
where $\sigma_{i}$ is the learnable parameter corresponding to the $i$-th loss term. $L_i$ represents the $i$-th loss term in ($L_{\mathrm{cross}}, L_{\mathrm{cls}}, L_{\mathrm{consist}}$). $\frac{1}{2\sigma_i^2}$ is used for dynamic weighting. $\log(1+\sigma_i^2)$ stands for a regularization term designed to constrain weight growth, thereby preventing any single loss term from being overemphasized.

\section{EXPERIMENTS}

\subsection{Dataset and Evaluation Metrics}
This study evaluates the proposed RDB method using two commonly utilized RSITR datasets: RSICD \cite{LU2018} and RSITMD \cite{YUAN20222}. 1) The RSICD dataset comprises 10,921 RS images gathered from various platforms such as Google Earth, Baidu Maps, MapABC, and Tianditu. All images are uniformly resized to a fixed dimension of 224×224 pixels. Each image is described by five texts with the same semantics but different expression forms. 2) The RSITMD dataset serves as a detailed and demanding benchmark for remote sensing image-text retrieval. Unlike traditional RS image-text datasets that focus on generic descriptions, RSITMD emphasizes the nuanced relationships between objects. In addition, each image is annotated with one to five keywords, supporting keyword-based RSITR tasks. RSITMD contains 4,743 images with size 256×256 pixels covering 32 distinct scenes, and it provides a total of 23,715 annotations.

To assess the model performance, two kinds of evaluation metrics are employed: Recall at K (R@K) with K values of 1, 5, and 10, and mean recall (mR). R@K measures the proportion of true matches that appear within the top K retrieval results, where K is set as 1, 5, and 10 in this study. Generally, R@K is used to evaluate the model accuracy at different retrieval depths. mR is calculated as the average of the R\@K scores obtained from both text-to-image and image-to-text retrieval tasks. It offers a comprehensive assessment of cross-modal retrieval performance. The specific calculation formula as follows:
\begin{equation}
    \mathrm{mR} = 
    (\underbrace{R@1 + R@5 + R@10}_{\text{Image-to-text retrieval}} +
    \underbrace{R@1 + R@5 + R@10}_{\text{Text-to-image retrieval}}) \,/\, 6
\end{equation}


\subsection{Implementation Details}
All experiments in this study were performed on NVIDIA RTX 4070 12GB GPUs. The dataset was partitioned following previous work \cite{YUAN20222}, with training, validation and test sets comprising 80\%, 10\%, and 10\% of the total data, respectively. Consistent with the GeoRSCLIP model, the final features representations were linearly projected into a 512-dimensional common space. Additionally, all images were resized to a fixed size of 224×224 pixels for training. In terms of parameter settings, both the dropout rate and $\mu$ in the cross-modal loss function were empirically set to 0.2. The initial learning rate of the network was set to 2e-4 and a weight decay of 0.7 was applied after every 20 training epochs. During the training process, the Adam optimizer was employed to update the network parameters. The batchsize was set to 32 and the total training epoch was set to 30. Finally, to ensure the stability and reliability of the experimental results, the experiments were evaluated using 5-fold cross-validation, with performance metrics reported as the average over all folds.

\subsection{Comparative Experiments}
In this study, the proposed method is compared with the state-of-the-art retrieval methods using two standard RSITR benchmark datasets. The experiments use GeoRSCLIP as the backbone network. Table~\ref{tab:results} highlights a comparison of retrieval results between traditional methods and CLIP-based methods on the RSICD and RSITMD datasets. From the table, it can be seen that the proposed RDB method improves the performance by 13\%-23\% over the traditional deep learning methods. CLIP-based PEFT methods leverage powerful generalization capabilities and extensive vision-language prior knowledge gained from pre-training on large-scale natural scene datasets, thereby achieving significant performance gains in RSITR tasks. Specifically, on the RSICD dataset, the proposed RDB method improves the mR by approximately 16.44\% compared to CLIP-Adapter and by 6.97\% compared to PE-RSITR. On RSITMD dataset, the proposed RDB method improves 13.98\% compared to Cross-Modal Adapter and 7.57\% compared to PE-RSITR. Notably, with the GeoRSCLIP pre-trained model, the proposed RDB method outperforms all current PEFT methods on both the RSICD and RSITMD datasets, and even surpasses the retrieval performance of the full fine-tuned GeoRSCLIP model. These improvements can be attributed to two factors: 1) The imbalanced cross-modal optimization problem is alleviated by designing modality-specific adapters for image and text modalities; 2) Multi-task joint optimization of cross-modal semantic alignment and intra-modal semantic discrimination facilitates cross-modal alignment robustness.

\begin{table*}[ht]
\centering
\caption{Retrieval performance on the RSICD and RSITMD test sets. \textcolor{red}{Red}: the proposed RDB method; \textcolor{blue}{Blue}: full fine-tuned GeoRSCLIP.}
\label{tab:results}
\resizebox{\textwidth}{!}{%
\begin{tabular}{lcccccccccc}
\toprule
\textbf{Methods} & \textbf{Backbone (image/text)} & \textbf{Tuned on} & & \multicolumn{3}{c}{\textbf{Image-to-text}} & \multicolumn{3}{c}{\textbf{Text-to-image}} & \textbf{mR} \\
&&& & \textbf{R@1} & \textbf{R@5} & \textbf{R@10} & \textbf{R@1} & \textbf{R@5} & \textbf{R@10} & \\
\midrule

\multicolumn{11}{l}{\textbf{Traditional methods}} \\
\midrule
AMFMN-soft~\cite{YUAN20222}   & ResNet18, biGRU & RSICD & & 5.05 & 14.53 & 21.57 & 5.05 & 19.74 & 31.04 & 16.02 \\
AMFMN-fusion~\cite{YUAN20222}   & ResNet18, biGRU & RSICD & & 5.39 & 15.08 & 23.40 & 4.90 & 18.28 & 31.44 & 16.42 \\
AMFMN-sim~\cite{YUAN20222}    & ResNet18, biGRU & RSICD & & 5.21 & 14.72 & 21.57 & 4.08 & 17.00 & 30.60 & 15.53 \\
GaLR w/o MR~\cite{YUAN2022}  & ResNet18, biGRU & RSICD & & 6.50 & 18.91 & 29.70 & 5.11 & 19.57 & 31.92 & 18.62 \\
GaLR with MR~\cite{YUAN2022} & ResNet18, biGRU & RSICD & & 6.59 & 19.85 & 31.04 & 4.69 & 19.48 & 32.13 & 18.96 \\
PIR~\cite{PAN2023}          & Swin Transformer, Bert & RSICD & & 9.88 & 27.26 & 39.16 & 6.97 & 24.56 & 38.92 & 24.46 \\
\midrule

\multicolumn{11}{l}{\textbf{CLIP-based methods}} \\
\midrule
Full-FT CLIP~\cite{pmlr-v139-radford21a}           & CLIP(ViT-B-32) & RSICD & & 15.89 & 36.14 & 47.93 & 12.21 & 32.97 & 48.84 & 32.33 \\
Full-FT GeoRSCLIP~\cite{ZHANG10679571}     & GeoRSCLIP(ViT-B-32-RET-2) & RSICD & & \textcolor{blue}{18.85} & \textcolor{blue}{38.15} & \textcolor{blue}{53.16} & \textcolor{blue}{14.27} & \textcolor{blue}{39.71} & \textcolor{blue}{57.49} & \textcolor{blue}{36.94} \\
Adapter~\cite{pmlr-v97-houlsby19a}               & CLIP(ViT-B-32) & RSICD & & 8.73 & 24.73 & 37.81 & 8.43 & 26.02 & 43.33 & 24.84 \\
CLIP-Adapter~\cite{GAO2023}          & CLIP(ViT-B-32) & RSICD & & 7.11 & 19.48 & 31.01 & 7.67 & 24.87 & 39.73 & 21.65 \\
AdaptFormer~\cite{CHEN2022}          & CLIP(ViT-B-32) & RSICD & & 12.46 & 28.49 & 41.86 & 9.09 & 29.89 & 46.81 & 28.10 \\
Cross-Modal Adapter~\cite{jiang2022cross}  & CLIP(ViT-B-32) & RSICD & & 11.18 & 27.31 & 40.62 & 9.57 & 30.74 & 48.36 & 27.96 \\
UniAdapter~\cite{lu2024uniadapter}          & CLIP(ViT-B-32) & RSICD & & 12.65 & 30.81 & 42.74 & 9.61 & 30.06 & 47.16 & 28.84 \\
PE-RSITR~\cite{YUAN2023}             & CLIP(ViT-B-32) & RSICD & & 14.13 & 31.51 & 44.78 & 11.63 & 33.92 & 50.73 & 31.12 \\
RSCLIP~\cite{cha2024pushing}               & ViT, Bert & RSICD & & 10.43 & 25.34 & 39.34 & 9.90 & 30.52 & 45.03 & 26.76 \\
\rowcolor[gray]{0.95}
RDB (Ours) & GeoRSCLIP(ViT-B-32-RET-2) & RSICD & & \textcolor{red}{18.66} & \textcolor{red}{41.08} & \textcolor{red}{53.89} & \textcolor{red}{15.70} & \textcolor{red}{41.17} & \textcolor{red}{58.02} & \textcolor{red}{38.09} \\
\midrule

\multicolumn{11}{l}{\textbf{Traditional methods}} \\
\midrule
AMFMN-soft~\cite{YUAN20222}   & ResNet18, biGRU & RSITMD & & 11.06 & 25.88 & 39.82 & 9.82 & 33.94 & 51.90 & 28.74 \\
AMFMN-fusion~\cite{YUAN20222}   & ResNet18, biGRU & RSITMD & & 11.06 & 29.20 & 38.72 & 9.96 & 34.03 & 52.96 & 29.32 \\
AMFMN-sim~\cite{YUAN20222}    & ResNet18, biGRU & RSITMD & & 10.63 & 24.78 & 41.81 & 11.51 & 34.69 & 54.87 & 29.72 \\
GaLR w/o MR~\cite{YUAN2022}  & ResNet18, biGRU & RSITMD & & 13.05 & 30.09 & 42.70 & 10.47 & 36.34 & 53.35 & 31.00 \\
GaLR with MR~\cite{YUAN2022} & ResNet18, biGRU & RSITMD & & 14.82 & 31.64 & 42.48 & 11.15 & 36.68 & 51.68 & 31.41 \\
PIR~\cite{PAN2023}          & Swin Transformer, Bert & RSITMD & & 18.14 & 41.15 & 52.88 & 12.17 & 41.68 & 63.41 & 38.24 \\
\midrule

\multicolumn{11}{l}{\textbf{CLIP-based methods}} \\
\midrule
Full-FT CLIP~\cite{pmlr-v139-radford21a}           & CLIP(ViT-B-32) & RSITMD & & 26.99 & 46.90 & 58.85 & 20.53 & 52.35 & 71.15 & 46.13 \\
Full-FT GeoRSCLIP~\cite{ZHANG10679571}     & GeoRSCLIP(ViT-B-32-RET-2) & RSITMD & & \textcolor{blue}{30.53} & \textcolor{blue}{49.78} & \textcolor{blue}{63.05} & \textcolor{blue}{24.91} & \textcolor{blue}{57.21} & \textcolor{blue}{75.35} & \textcolor{blue}{50.14} \\
CLIP-Adapter~\cite{GAO2023}          & CLIP(ViT-B-32) & RSITMD & & 12.83 & 28.84 & 39.05 & 13.30 & 40.20 & 60.06 & 32.38 \\
AdaptFormer~\cite{CHEN2022}          & CLIP(ViT-B-32) & RSITMD & & 16.71 & 30.16 & 42.91 & 14.27 & 41.53 & 61.46 & 34.81 \\
Cross-Modal Adapter~\cite{jiang2022cross}  & CLIP(ViT-B-32) & RSITMD & & 18.16 & 36.08 & 48.72 & 16.31 & 44.33 & 64.75 & 38.06 \\
UniAdapter~\cite{lu2024uniadapter}          & CLIP(ViT-B-32) & RSITMD & & 19.86 & 36.32 & 51.28 & 17.54 & 44.89 & 56.46 & 39.23 \\
PE-RSITR~\cite{YUAN2023}             & CLIP(ViT-B-32) & RSITMD & & 23.67 & 44.07 & 60.36 & 20.10 & 50.63 & 67.97 & 44.47 \\
RSCLIP~\cite{cha2024pushing}               & ViT, Bert & RSITMD & & 19.25 & 36.06 & 46.68 & 12.92 & 42.04 & 63.14 & 36.68 \\
\rowcolor[gray]{0.95}
RDB (Ours) & GeoRSCLIP(ViT-B-32-RET-2) & RSITMD & & \textcolor{red}{32.08} & \textcolor{red}{53.98} & \textcolor{red}{67.26} & \textcolor{red}{25.27} & \textcolor{red}{58.58} & \textcolor{red}{75.09} & \textcolor{red}{52.04} \\
\bottomrule
\end{tabular}%
}
\end{table*}

\subsection{Ablation Study}
To validate the effectiveness of the proposed RDB method, this study compares it against three baseline methods: 1) the full fine-tuned GeoRSCLIP method, 2) the GeoRSCLIP method using CMAA but without introducing DTCL, and 3) the GeoRSCLIP method using both CMAA and DTCL. All experiments were performed using the RSITMD dataset, and Table~\ref{tab:ablas1} summarizes the results of the ablation study.

\begin{table*}
\centering
\caption{Performance comparison of the three baseline methods on the RSITMD dataset.}
\label{tab:ablas1}
\resizebox{\textwidth}{!}{%
\begin{tabular}{c|c|cc|ccc|ccc|c}
\toprule
\textbf{Method} & \textbf{Backbone} & \multicolumn{2}{c|}{\textbf{Module}}
& \multicolumn{3}{c|}{\textbf{Image-to-text}} 
& \multicolumn{3}{c|}{\textbf{Text-to-image}} 
& \textbf{mR} \\
 &  & CMAA & DTCL & R@1 & R@5 & R@10 & R@1 & R@5 & R@10 & \\
\hline
Full-FT GeoRSCLIP & ViT-B-32-RET-2 & \ding{55} & \ding{55} & 30.53 & 49.78 & 63.05 & 24.91 & 57.21 & 75.35 & 50.14 \\
\hline
GeoRSCLIP w/CMAA & ViT-B-32-RET-2 & \ding{51} & \ding{55} & 30.80 & 50.88 & 66.81 & 25.07 & 58.36 & 75.44 & 51.23 \\
GeoRSCLIP (RDB) & ViT-B-32-RET-2 & \ding{51} & \ding{51} & 32.08 & 53.98 & 67.26 & 25.27 & 58.58 & 75.09 & 52.04 \\
\bottomrule
\end{tabular}%
}
\end{table*}

In this study, GeoRSCLIP (ViT-B-32-RET-2) is employed as the backbone model for the ablation experiments. As shown in Table~\ref{tab:ablas1}, the first row presents the fully fine-tuned GeoRSCLIP result on the downstream dataset, which achieves an mR of 50.14. Although this performance is promising, full fine-tuning requires updating all model parameters, leading to high computational cost and limited scalability. In contrast, fine-tuning with CMAA (second row) improves performance, yielding an mR of 51.23. Moreover, incorporating DTCL alongside CMAA (third row) further enhances the performance to an mR of 52.04. These experimental results validate the effectiveness of these modules and demonstrate that proposed RDB method effectively addresses the problem of representation discrepancy between modalities.

To evaluate the impact of each DTCL component on the overall model performance, an ablation study is conducted on its the three loss terms. The experimental results are shown in Table~\ref{tab:ablas2}. Method 1 presents the results of introducing only cross-modal constraint (GeoRSCLIP w/ cross loss). Method 2 presents the results of introducing both cross-modal and classification constraints (GeoRSCLIP w/ cross loss + cls loss). Method 3 presents the result by combining cross-modal and consistency constraints (GeoRSCLIP w/ cross loss + consist loss). The fourth row is the result of introducing cross-modal, classification, and consistency constraints simultaneously (GeoRSCLIP(RDB)).

\begin{table*}
\centering
\caption{Performance comparison of different loss term combinations in DTCL.}
\label{tab:ablas2}
\resizebox{\textwidth}{!}{%
\begin{tabular}{c|ccc|ccc|ccc|c}
\toprule
\textbf{Method} & \multicolumn{3}{c|}{\textbf{Loss}}
& \multicolumn{3}{c|}{\textbf{Image-to-text}} 
& \multicolumn{3}{c|}{\textbf{Text-to-image}} 
& \textbf{mR} \\
 & cross & cls & consist & R@1 & R@5 & R@10 & R@1 & R@5 & R@10 & \\
\hline
Method 1 & \ding{51} & \ding{55} & \ding{55} & 30.80 & 50.88 & 66.81 & 25.07 & 58.36 & 75.44 & 51.23 \\
Method 2 & \ding{51} & \ding{51} & \ding{55} & 30.97 & 	52.43 & 67.26 & 25.49 & 58.32 & 75.62 & 51.68 \\
Method 3 & \ding{51} & \ding{55} & \ding{51} & 30.97 & 51.99 & 66.37 & 25.31 & 58.36 & 75.58 & 51.43 \\
GeoRSCLIP (RDB) & \ding{51} & \ding{51} & \ding{51} & 32.08 & 53.98 & 67.26 & 25.27 & 58.58 & 75.09 & 52.04 \\
\bottomrule
\end{tabular}%
}
\end{table*}

From the results in Table~\ref{tab:ablas2}, it can be seen that when only the cross-modal constraint is employed, the model achieves a mean Recall (mR) of 51.23. When the cross-modal constraint is combined individually with the classification constraint and the consistency constraint, the mR improves to 51.68 and 51.43, respectively. This demonstrates the effectiveness of extending single-task optimization to multi-task optimization. Further, when cross-modal, classification, and consistency constraints are introduced simultaneously, the model performance is significantly improved (52.04 vs 51.23). This demonstrates the substantial benefit of extending single-task optimization to a multi-task framework via DTCL.

\subsection{Qualitative Analysis}

\subsubsection{Image-Text Results Analysis}
\begin{figure*}[tp]
\begin{center}
\includegraphics[width=1.0\linewidth]{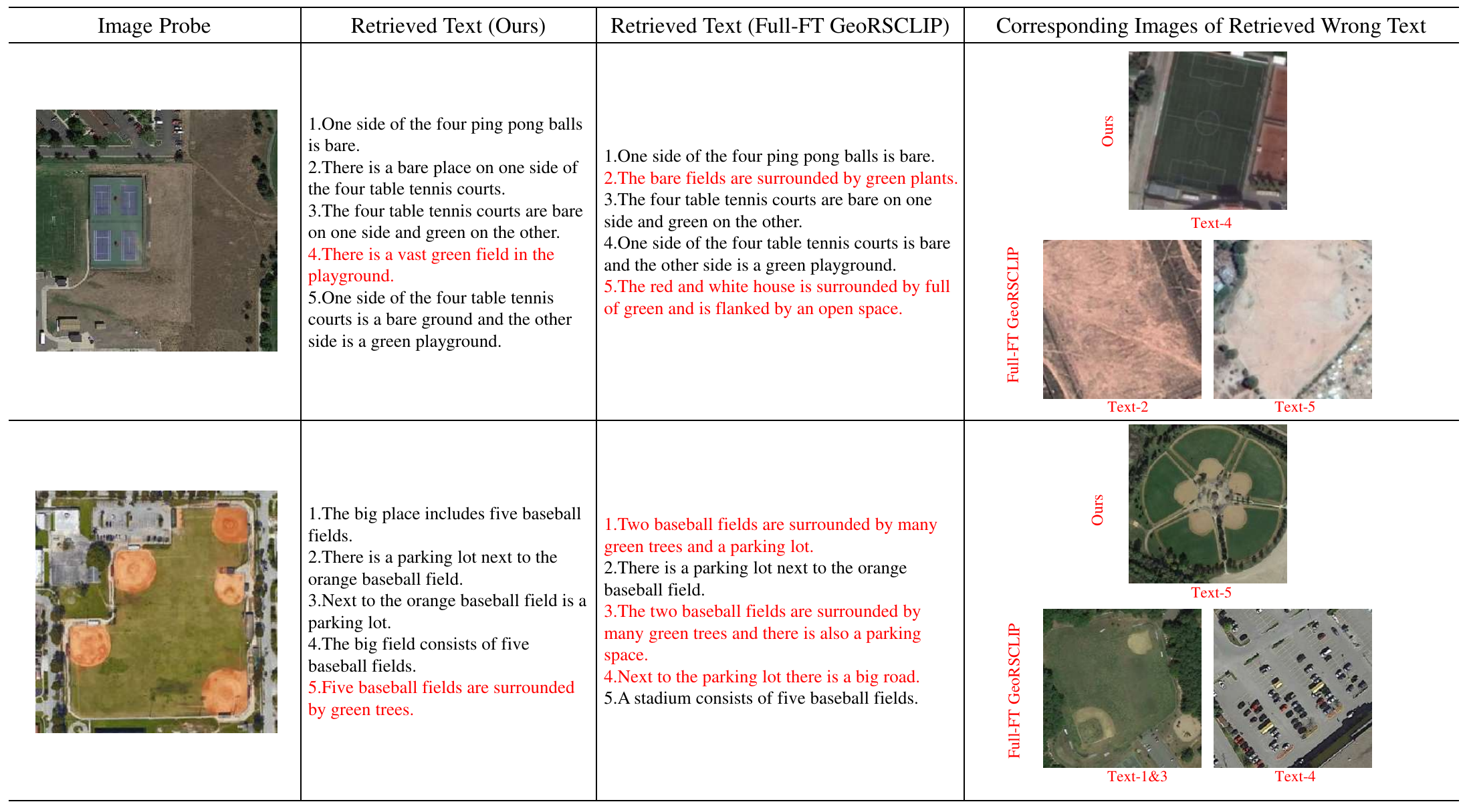}
\renewcommand{\figurename}{Fig.}
\end{center}
    \caption{\small{Comparison of Top-5 results between the proposed RDB method and the Full-FT GeoRSCLIP method on the RSITMD dataset for the image-text retrieval task. Red markings indicate retrieval errors; the last column shows the correct RS image corresponding to the incorrectly retrieved text.}}
\label{fig:irt}
\end{figure*}

Figure~\ref{fig:irt} compares the Top-5 image-text retrieval results of the proposed RDB method with Full-FT GeoRSCLIP on the RSITMD dataset. From the results, it can be seen that the proposed RDB method outperforms Full-FT GeoRSCLIP in terms of retrieval performance. Although not all the Top-5 retrieval results of the proposed RDB method are entirely correct, its error results maintain strong semantic consistency with the main goal described by ground truth. For example, in the first image, although the Text-4 retrieved by the proposed RDB method exhibits some minor deviations, its corresponding ground truth is “playground”. In contrast, the Full-FT GeoRSCLIP predominantly focuses on the expansive “bareland” regions, resulting in the ground truths for Text-2 and Text-5 being labeled as “bareland” rather than “playground”. In the second image, although the proposed RDB method produced an erroneous retrieval, the retrieved text still accurately identifies the number of “baseball field”. In contrast, the retrieved Text-1 and Text-3 by Full-FT GeoRSCLIP only focus on the primary objects, but fails to capture the quantitative information. Meanwhile, the retrieved Text-4 by Full-FT GeoRSCLIP mistakenly identifies “parking” as the key object in the image. The above phenomenon indicates significant differences between the proposed RDB method and Full-FT GeoRSCLIP when extracting image features in complex RS scenes. Full-FT GeoRSCLIP fails to capture certain key details during the cross-modal alignment. It leads to discrepancies between the text modality retrieved via the image probe and the ground truth. In contrast, the proposed RDB method demonstrates a superior ability in fine-grained image information learning, which further validates the superiority of the proposed RDB method in cross-modal retrieval tasks.

\subsubsection{Text-Image Results Analysis}
\begin{figure*}[tp]
\begin{center}
\includegraphics[width=1.0\linewidth]{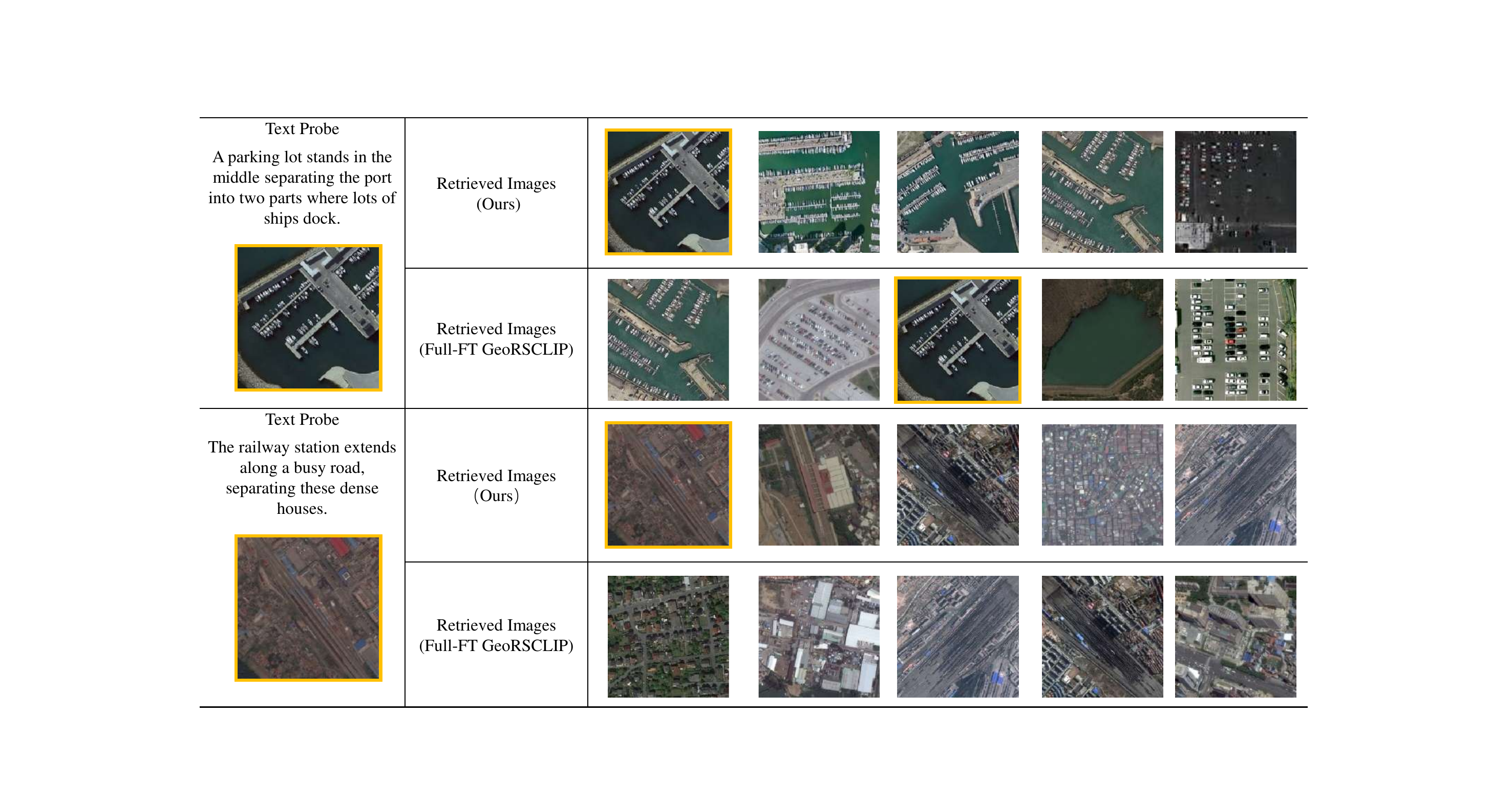}
\renewcommand{\figurename}{Fig.}
\end{center}
    \caption{\small{Comparison of Top-5 results between the proposed RDB method and the Full-FT GeoRSCLIP method in the text-image retrieval task on the RSITMD dataset. The portion marked by the orange box indicates the RS image that matches the retrieved text.}}
\label{fig:tri}
\end{figure*}

Figure~\ref{fig:tri} compares the Top-5 text-image retrieval results of the proposed RDB method with Full-FT GeoRSCLIP on the RSITMD dataset. The comparative analysis shows that the proposed RDB method significantly outperforms Full-FT GeoRSCLIP in terms of retrieval performance. Specifically, for the first text, the key object is “port”. The proposed RDB method can accurately identify the corresponding image at the Top-1 rank. In addition, the retrieval results from Top-2 to Top-4 also correctly correspond to “port”. It demonstrates that the proposed RDB method can effectively capture the primary semantic object described in the text. However, since the text also mentions “parking”, the Top-5 result mistakenly identifies the key object as “parking”. In contrast, Full-FT GeoRSCLIP only retrieved correct results in Top-3. Although the ground truth of Top-1 is also “port”, the model fails to capture the fine-grained information in the text. Top-2 and Top-5 misidentify the key object as “parking”, while Top-4 result is entirely unrelated to the textual information. The retrieval results of the second text are similar to the first one. The proposed RDB method successfully retrieved an accurate match at Top-1. The results at Top-2, Top-3 and Top-5 are also consistent with the ground truth of the text probe, except that Top-4 misidentifies the key object as “dense residential”. In contrast, the Top-5 results of Full-FT GeoRSCLIP fail to retrieve the correct RS images. Although some retrieved images match the ground truth of the text probe, they fail to capture the fine-grained information, leading to a significant bias in the retrieval results.

These results fully demonstrate the advantages of the proposed RDB method in fine-grained feature extraction and cross-modal information alignment. The method effectively bridges the differences between image and text modalities. It also significantly improves the retrieval performance and generalization ability of the model.

\section{CONCLUSION}
In this study, a Representation Discrepancy Bridging (RDB) method is proposed for addressing the imbalanced cross-modal optimization problem in the RSITR task. On the one hand, a Cross-Modal Asymmetric Adapter (CMAA) is developed to optimize modality-specific features for images and text, respectively. On the other hand, a DTCL loss function is designed for enhancing cross-modal semantic alignment by adaptively weighting cross-modal constraints, classification constraints, and exponential moving average consistency constraints. Experimental results conducted on the RSICD and RSITMD datasets confirm the effectiveness of CMAA and DTCL loss. In addition, experimental results shows the proposed RDB method achieves substantial performance improvements and even outperforms full fine-tuning methods. Although the proposed RDB method performs well in the RS cross-modal retrieval task, the semantic diversity in complex RS images may lead to incomplete coverage of text descriptions and exacerbate inter-modal information asymmetry. Future work will introduce generative text enhancement techniques to extend the semantic coverage of text descriptions and further improve retrieval performance in complex scenarios.

\section*{ACKNOWLEDGMENTS}
This work was supported in part by National Natural Science Foundation of China under Grant 62201452 and Grant 62271296, in part by Technology Innovation Guidance Special Fund of Shaanxi Province under Grant 2024QY-SZX-17, and in part by The Innovation Capability Support Plan Project in Shaanxi Province under Grant 2025RS-CXTD-012.

\bibliography{mybibfile}

\end{document}